\documentclass[11pt,a4paper]{article}

\usepackage[utf8]{inputenc}
\usepackage[T1]{fontenc}
\usepackage[english]{babel}
\usepackage{amsmath,amssymb}
\usepackage{hyperref}
\usepackage{url}
\usepackage{graphicx}
\usepackage{microtype}
\usepackage{textcomp}
\usepackage{booktabs}
\usepackage[margin=2.5cm]{geometry}
\hypersetup{
    pdftitle={On the missing benchmarks layer and a potential solution},
    pdfauthor={Francis F Daniel, Mauro Ibañez, Francis Perelman, Marian Basti},
    pdfkeywords={AI, Benchmarks, Latin America, Evals Hub, multipolar AI},
    colorlinks=true,
    linkcolor=blue,
    citecolor=blue,
    urlcolor=blue
}

\title{On the missing benchmarks layer and a potential solution}
\author{
  Francis F Daniel$^*$\textsuperscript{1} \and
  Mauro Ibañez$^*$\textsuperscript{1} \and
  Francis Perelman\textsuperscript{2} \and
  Marian Basti\textsuperscript{1}
}
\date{July 29, 2026}

\begin{document}
\maketitle

\begin{center}
\textsuperscript{1}SURUS \quad \textsuperscript{2}Independent\\[0.3em]
$^*$Equal contribution (co-first authors)\\[0.2em]
Contact: \texttt{francis@surus.lat}
\end{center}

\begin{abstract}
Latin America is missing a foundational layer for native AI development: the benchmark layer. The benchmark layer does two things no other layer can — it audits AI systems against regional social requirements and it directs AI optimization in economically relevant environments. Without it, public institutions cannot independently evaluate foreign AI systems, and companies cannot optimize AI systems to solve local problems with SOTA performance. The cost of the missing layer is dual: a loss of auditability and a loss of optimization direction over a technology that is increasingly critical infrastructure. We propose an EvalsHub, with LatamBoard as its first regional instance: an open, task-first benchmark infrastructure where universities, public institutions, professional communities, and companies can publish, execute, compare, and maintain evaluations across models, workflows, and agents. Built once, measured forever — re-run by institutions as new AI systems ship and by industry teams after every system change. Open by design and incentive-driven by construction.
\end{abstract}


\tableofcontents
\newpage

\section{The missing benchmark layer}

\subsection{Frame}

Latin America is missing a foundational layer for native AI development --- the benchmark layer.

\subsection{Public-institution side of the absence}

On the public-institution side, almost every AI system in regional use today was built elsewhere, and almost none of it has been measured against the contexts it is being deployed into. In practice, AI procured by public-institutions arrives as a foreign-built artifact and cannot be determined if it is appropriate for the desired use and if has the right value-system. In the absence of a regionally grounded benchmark layer, institutions default to vendor claims and have no instrument to evaluate the AI they consume \cite{metaxa2021auditing}.

\subsection{Industry side of the absence}

On the industry side, modern AI optimization --- the software 3.0 stack of prompt-program optimizers and workflow-architecture search \cite{reynolds2021prompt} --- consumes a benchmark as its input; no benchmark means no optimization target. Software 3.0-stack optimizers take a benchmark and a system as inputs and search for prompt programs or workflow-architecture changes that produce a higher score against that benchmark. No benchmark means no target, the optimizer cannot run, and the general-purpose model remains at its out-of-box performance on tasks with a relevant regional context. For example, a pest-classification model trained on European or North American agricultural imagery will not reliably identify the species affecting a Colombian coffee crop. The model appears to work and returns plausible output, but without a benchmark the miscalls remain invisible on exactly the tasks the regional economy needs AI to perform best \cite{manvi2024geobias}.

\subsection{The dual cost}

The cost of this missing layer is dual: a loss of auditability on one side, a loss of optimization direction on the other, and together a loss of industrial productivity and sovereign capacity over increasingly critical infrastructure. In practical terms, the region cannot audit what it consumes and cannot direct what it builds, cutting off both hands from a technology that is increasingly critical infrastructure \cite{timmers2019sovereignty}. The next section names what the benchmark layer does to close both gaps --- the audit function on the public-institution side and the optimization target on the industry side.

\section{What the benchmark layer does: audit and direction}

\subsection{Frame}

The benchmark layer does two things no other layer can.

\subsection{Function one: the audit function}

On the public-institution side, benchmarks reveal whether a foreign or regional AI model acts correctly in a given context. The audit function is the principal lever public institutions have for understanding and governing AI they did not build \cite{liu2022medicalaudit}. A benchmark measures behavior on the tasks and contexts (domains and languages) that matter, and produces a score that public institutions can feed into procurement, policy, and oversight decisions. This can position public institutions as independent auditors of AI systems worldwide, the same way some universities independently measure poverty or inequality \cite{mokander2023auditing}.

\subsection{Function two: the optimization function}

On the industry side, software 3.0-stack optimizers (e.g., DSPy \cite{khattab2023dspy} with MIPRO/GEPA-style prompt search \cite{opsahlong2024mipro, agrawal2025gepa}) take a benchmark and output a more performing and/or aligned AI system, allowing a general-purpose model to be brought to state-of-the-art performance on a regional task through prompt, workflow-architecture, and inference-parameter changes --- with no retraining and no inference-infrastructure change. Concretely, an optimizer is a program that consumes the benchmark and a starting AI system and searches for prompt-program, workflow-architecture, and inference-parameter changes that produce a higher score --- with no model-parameter retraining and no change to the inference infrastructure. This is how an AI team at a company brings a general-purpose model, foreign or regional, to state-of-the-art performance on the specific task and context they care about, without training and owning a foundation model \cite{bommasani2021foundationmodels}.

\subsection{One artifact, two consumers, two functions}

One artifact, two consumers, two functions --- a public good for institutions and an optimization compass for industry. For institutions, benchmarks published openly can be run by any organization; the audit function scales because the artifact is shared, not rebuilt per body. For industry, a team pulls the same benchmark as the optimization target for a private production system and re-runs it after every change for continuous performance tracking. The same artifact serves both consumers --- one authoring effort, two consumption paths, no duplication --- which is what makes the benchmark layer shared infrastructure rather than N independent evaluations \cite{liang2022helm}.

\section{An EvalsHub and LatamBoard: a proposed solution}

\subsection{Proposal and instance}

We propose a regional EvalsHub, with LatamBoard ([latamboard.ai](https://latamboard.ai)) as the first instance --- the Latin American surface where benchmarks built by and for the region are published, runnable, and comparable across AI systems. It is the common web surface where regional benchmarks are published, indexed, and their results compared across AI models (e.g. DeepSeek vs GPT vs LatamGPT \cite{latamgpt2025}). Every benchmark on the hub is published under an open license, runnable end-to-end with reproducible execution, and comparable via a standardized evaluation protocol so scores carry across AI system, be it a model, node, workflow or agent.

\subsection{Task-first ontology}

\texttt{/<task?>/<domain?>/<language?>} names a task-first ontology: task first because a benchmark is an exam for something; domain second because task context matters; language third as a modifier, with each level modifying the last. A benchmark is an exam, and every exam is for a specific capability --- e.g., extraction, classification, summarization, transcription, translation, reasoning, etc. Domain sits second because "correct" depends on context --- medical classification $\neq$ legal classification $\neq$ finance classification. Language sits third because regional language varieties matter --- Chilean Spanish differs from Colombian Spanish \cite{goncalves2015dialects}. Each level compounds the last so the artifact at, for example, \texttt{/extract/medical/es-CL} is precise about all three axes.

\subsection{AI-system agnostic}

The approach is AI-system agnostic: the benchmark defines what is being tested, while the AI target (model, node, workflow, agent) defines who is being tested. What stays fixed is the exam itself --- the task definition, inputs, reference answers, and scoring function. What varies is the AI system under test --- e.g., a single model call, a node inside a workflow, a full workflow, or an agent that orchestrates multiple steps. This matters because modern AI in production is compound, and a legal-document summarization model trained primarily on US case law may miss the procedural structure of an Argentine civil filing unless the benchmark measures the system actually deployed, not just an isolated model \cite{bender2021stochastic}.

\subsection{Built once, measured forever}

Every benchmark can be built once, and measured forever --- that is, re-run by an organization each time a new AI model is created (foreign or regional), and re-run by an industry team after every change to the system being optimized (for continuous performance tracking). On the institutional side, every time a new model or version is created, re-running the same benchmark compounds the artifact's value as scores accumulate across systems and time, and it makes otherwise plausible outputs legible as shifts rather than silent failures. On the industry side, each workflow or prompt change is followed by a re-run on the same benchmark to get the new score --- this is continuous performance tracking and is what a proper optimization target enables. Authoring is one-time while evaluation is forever, which is what makes a benchmark a piece of infrastructure rather than a project. An open question is how to guard against benchmark data contamination as models increasingly train on web-scale corpora that may include benchmark data \cite{xu2024contamination}.

\section{The Access Problem: the binding constraint on regional benchmark supply}

\subsection{Frame}

The benchmark layer faces a hard supply problem --- benchmarks must be authored from scratch, by humans. Datasets can be compiled from records that already exist (e.g., transactions, logs, corpora, publications), but a benchmark is an exam that must be designed and written. It encodes a claim about what \textit{correct} looks like for a task in a given domain, and only a domain authority can make that claim \cite{weber2019benchmarking}.

\subsection{The four expertises required}

Authoring a single benchmark today requires four kinds of expertise that almost never sit in the same person: knowledge of what \textit{correct} means in a domain (the domain expert), the ability to write a benchmark schema and scoring function (machine learning engineering), the theory and practice of generating a representative test (statistics), and the operational know-how to run the resulting artifact reliably (developer) \cite{hutchinson2022evalgaps}. A physician (or historian, agronomist, lawyer) knows what \textit{correct} means in their domain; an ML engineer knows how to encode that judgment as a schema and scoring function; a statistician knows how to draw a representative test; a developer knows how to ship the artifact so it runs reliably. A benchmark for medical named-entity recognition in Mexican health records, for example, needs a practicing clinician's ground truth \cite{schneider2020biobertpt}, an ML engineer's encoding of the test schema and its scoring function, a statistician's benchmark data distribution, and a developer's packaging and run automation. Only a very small fraction of professionals have these 4 kinds of expertise, and thus a very small percentage of people can author a benchmark end-to-end.

\subsection{The Access Problem}

The people whose judgment is most valuable to the region --- domain experts --- are locked out by the other three, a situation we call the Access Problem and the binding constraint on regional benchmark supply \cite{kay2024epistemic}. Their judgment is the most valuable because they carry the substantive claim about what \textit{correct} looks like for the regional task; without that claim a benchmark can be technically well-formed yet epistemically empty. The other three expertises --- ML engineering, statistics, and developer operations --- are the price of entry, and when domain experts lack them their judgment does not reach the artifact. The consequence is structural: no amount of money, compute, or attention poured into the other three unlocks supply if domain experts remain locked out, so for EvalsHub to reach scale, it must be paired with a technology that lowers the ML-engineering, statistics, and developer-operations barrier of entry so domain experts can author benchmarks directly. This is still an open problem.

\section{A Multipolar position: Open by design, incentive-driven by construction}

\subsection{Stance}

We prefer a multipolar AI world --- many regions, many communities, many value systems --- and Latin America should be one of those poles, with its own technology and its own agency. The alternative is a unipolar/bipolar world in which the values, examples, tasks, and priorities of one or two regions dominate by default, with most widely used AI models and benchmarks originating there. Owning technology and agency means, at minimum, being able to steer imported models toward regional-context aware tasks and to audit them against regional standards. The development of sovereign foundation AI models is much more desirable \cite{roberts2024sovereignty}, and none of these two can be acheived without a benchmark layer.

\subsection{Design consequence}

From that position, an EvalsHub like LatamBoard must be open by design --- to foster the collaboration and innovation the region needs to reach the AI frontier --- and incentive-driven by construction --- to keep the evaluation layer as public infrastructure sustained continuously by the broader community, rather than a single team's project. Open by design means open licensing of the benchmark artifact, open source of the scoring code, and open access to the results (e.g., leaderboards visible), or the layer walls off collaboration and cannot compound with the region's community effort \cite{reuel2024betterbench}. Incentive-driven by construction means contribution rewards the contributor in a meaningful way (e.g. recognition, brand visibility) so as to keep a reinforcing loop \cite{lernertirole2002opensource}.

\subsection{Contributor recognition and reciprocity}

Contributor recognition and reciprocity are explicit: universities, public institutions, professional communities, and companies willing to open-source internal benchmarks are named as contributors to the region's evaluation layer, gain brand awareness on a heavy topic, and also indirectly benefit from benchmarks contributed by other organizations that are now more incentiviced to contribute as well. Universities gain institutional visibility on a heavy topic, public institutions gain soft-power positioning, professional communities gain a shared instrument for the field, and companies willing to open-source internal benchmarks gain a an understanding of the best performing AI systems worldwide for the tasks and context they care about.

\section{Open questions}

\subsection{Frame}

Many questions remain open.

\subsection{Methodological standards \& Logistics}

Methodological standards for a regional evaluation layer need to be defined and shared. Shared standards must cover how to design a benchmark schema, how build statistically representative benchmark data, how to score it (e.g., deterministic metrics, LLM-as-judge \cite{zheng2023llmjudge}, or human raters), and how to version and update the artifact as understanding of the task evolves \cite{pineau2020reproducibility}. This article does not attempt to close these questions --- it opens them for the region's community to converge on.

How often a benchmark should be re-run as new models ship, who maintains the artifacts, who pays for the compute, etc. --- has no shared answer yet.

\subsection{Sensitive-domain governance}

Sensitive-domain governance for benchmarks in domains like healthcare requires institutional structures that do not yet exist regionally. In these domains, a benchmark must carry the authority of the profession to shape procurement and practice \cite{reddy2019governance}. Regionally, the institutions that could confer that authority  have not yet organized around this role \cite{alami2026governance}.

\section{An invitation to contribute}

The benchmark layer only exists to the extent that the region's community builds it. LatamBoard is a starting point --- the surface, the ontology, and the first regional benchmarks --- but the layer itself is what universities, public institutions, professional communities, and companies choose to contribute over time.

We invite each of these stakeholders directly. Universities and research groups can publish the benchmarks they already build in the course of their research. Public institutions can contribute the evaluations they run --- or want to run --- on the AI they procure and oversee. Professional communities can claim authorship over the benchmarks that define correctness for their domain. Companies willing to open-source internal evaluations can turn private measurement work into public infrastructure, and gain a comparable view of AI performance across the tasks they care about \cite{franquesa2017sustainability}.

The ambition of LatamBoard is to become common infrastructure for AI development in the region --- built by the region, for the region. Every added benchmark makes the layer more valuable to every other participant, and every stakeholder that joins compounds the incentive for the next. Come find us at [latamboard.ai](https://latamboard.ai).

\bibliographystyle{plain}
\bibliography{main}

\end{document}